\documentclass[sigconf]{acmart}

\usepackage{graphicx}
\usepackage{amsmath, amssymb}
\usepackage{booktabs}
\usepackage{xcolor}
\usepackage{algorithm}
\usepackage{algorithmic}
\usepackage{subcaption}

\begin{document}

\title{Lost in OCR Translation? Vision-Based Approaches to Robust Document Retrieval}

\author{Alexander Most}
\email{AMost@lanl.gov}
\affiliation{%
  \institution{Los Alamos National Laboratory}
  \city{Los Alamos}
  \state{NM}
  \country{USA}
}

\author{Joseph Winjum}
\email{Joseph.Winjum@gmail.com}
\affiliation{%
  \department{Gianforte School of Computing}
  \institution{Montana State University}
  \city{Bozeman}
  \state{MT}
  \country{USA}
}

\author{Ayan Biswas}
\email{Ayan@lanl.gov}
\affiliation{%
  \institution{Los Alamos National Laboratory}
  \city{Los Alamos}
  \state{NM}
  \country{USA}
}

\author{Shawn Jones}
\email{SMJones@lanl.gov}
\affiliation{%
  \institution{Los Alamos National Laboratory}
  \city{Los Alamos}
  \state{NM}
  \country{USA}
}

\author{Nishath Rajiv Ranasinghe}
\email{Ranasinghe@lanl.gov}
\affiliation{%
  \institution{Los Alamos National Laboratory}
  \city{Los Alamos}
  \state{NM}
  \country{USA}
}

\author{Dan O'Malley}
\email{OMalleD@lanl.gov}
\affiliation{%
  \institution{Los Alamos National Laboratory}
  \city{Los Alamos}
  \state{NM}
  \country{USA}
}

\author{Manish Bhattarai}
\email{Ceodspspectrum@lanl.gov}
\affiliation{%
  \institution{Los Alamos National Laboratory}
  \city{Los Alamos}
  \state{NM}
  \country{USA}
}

\acmConference[DocEng '25]{ACM Symposium on Document Engineering}{August 2025}{Location TBD}

\begin{abstract}
Retrieval-Augmented Generation (RAG) has become a popular technique for enhancing the reliability and utility of Large Language Models (LLMs) by grounding responses in external documents. Traditional RAG systems rely on Optical Character Recognition (OCR) to first process scanned documents into text. However, even state-of-the-art OCRs can introduce errors, especially in degraded or complex documents. Recent vision-language approaches, such as ColPali, propose direct visual embedding of documents, eliminating the need for OCR. This study presents a systematic comparison between a vision-based RAG system (ColPali) and more traditional OCR-based pipelines utilizing Llama 3.2 (90B) and Nougat OCR across varying document qualities. Beyond conventional retrieval accuracy metrics, we introduce a semantic answer evaluation benchmark to assess end-to-end question-answering performance. Our findings indicate that while vision-based RAG performs well on documents it has been fine-tuned on, OCR-based RAG is better able to generalize to unseen documents of varying quality.
We highlight the key trade-offs between computational efficiency and semantic accuracy, offering practical guidance for RAG practitioners in selecting between OCR-dependent and vision-based document retrieval systems in production environments. 
\end{abstract}

\maketitle

\section{Introduction}
Large Language Models (LLMs) have shown improvements across the landscape of natural language processing, enabling promising performance increases across a multitude of tasks such as semantic analysis, question answering, machine translation, and text descriptions. However, despite their strengths, these models suffer from fundamental limitations due to their reliance on static training corpora, often leading to hallucinations or outdated information. Retrieval-Augmented Generation (RAG) addresses this limitation by allowing LLMs to retrieve and cite information from external sources during inference \cite{lewis2020retrieval}. In a typical RAG pipeline, user queries are transformed and stored as vector embeddings. The most relevant documents are then passed into the LLM’s context window along with the original query to guide generation. Leveraging this approach, RAG has been leveraged to boost code-translation quality ~\cite{bhattarai_RAG}\cite{bhattarai-etal-2025-enhancing} and has likewise been shown to curb hallucinations ~\cite{bhattarai-etal-2025-heal}.

In addition to improving response accuracy, the retrieval component of RAG substantially facilitates the application of LLMs to proprietary and evolving document collections. Unlike traditional fine-tuning methods—which are costly, time-consuming, and must be repeated whenever new documents are introduced—RAG enables real-time integration of the latest information through on-demand retrieval. Consequently, RAG has been widely adopted in commercial production settings, powering proprietary systems like ChatGPT, Claude, Perplexity, and Gemini, although detailed technical implementations remain largely unpublished in peer-reviewed venues.

This work specifically addresses the application of RAG to scanned or digitized documents. Traditionally, handling such documents involves first extracting text using OCR algorithms prior to embedding and retrieval. However, OCR preprocessing frequently introduces significant noise, especially in scenarios involving low-quality images, handwritten annotations, or complex layouts. Recent VLMs, such as ColPali~\citep{faysse2024colpali}, offer a promising alternative by directly embedding document images into unified multimodal vector spaces without OCR. These vision-based methods inherently preserve critical spatial relationships, formatting cues, and visual nuances that OCR processes often overlook or misinterpret.

Despite the potential advantages of direct image embedding, critical gaps persist in the current literature. Existing benchmarks for document retrieval systems, such as ViDoRe~\citep{faysse2024colpali}, predominantly evaluate performance using clean, high-quality documents, conditions rarely representative of practical real-world scenarios. Additionally, prior comparative evaluations—including those involving ColPali—have often relied on outdated OCR baselines like Tesseract~\citep{TessOverview} and have limited their analyses primarily to retrieval accuracy, neglecting comprehensive semantic question-answering capabilities. To date, the field lacks rigorous comparative assessments of OCR-to-RAG versus VLM-to-RAG pipelines across varied document quality conditions, particularly emphasizing downstream semantic accuracy and practical robustness.

Moreover, existing vision-language models suitable for direct embedding (e.g., ColPali, ColQwen, ColSmol) typically employ relatively lightweight architectures with parameter counts ranging between 2–7 billion, whereas current OCR-to-RAG pipelines can leverage significantly larger language models. Specifically, in this study, our OCR-based pipeline integrates Llama 3.2 (90 billion parameters), potentially affording substantial advantages in downstream semantic quality.

While one could theoretically adapt a 90B parameter model for the VLM role in ColPali, implementing this architecture presents significant practical challenges related to performance and memory usage. ColPali relies on generating and storing multiple dense vector embeddings per document (representing image patches), and using a VLM of 90B magnitude would dramatically increase the computational time needed to generate these numerous embeddings. Additionally, the memory and disk space required to store these multi-vector representations would likely become prohibitive for substantial document collections. The subsequent question embedding before the retrieval step would also face substantial slowdowns, potentially rendering the RAG system unusable for interactive applications. Although an operation might tolerate the costs associated with document embedding, the degraded query response times would likely make the system unacceptable to end users. 

Our empirical results show that even when using the same query encoder (Qwen2 7B), the VLM-based pipeline exhibits slower query times than the OCR-based system. This slowdown stems from the need to compare the query embedding against many patch-level image embeddings per document during late-interaction retrieval, making latency a practical bottleneck even with mid-sized models.

Given the increase in retrieval speed inherent in scaling the VLM component within this specific dense retrieval architecture, this study focuses on comparing representative, readily deployable model sizes characteristic of each paradigm (lightweight VLM vs. OCR + large LLM). This allows for an evaluation of the trade-offs presented by systems configurations commonly encountered in practice.

This study rigorously compares a VLM-based RAG pipeline (ColPali) against a competitive OCR-based pipeline utilizing Llama 3.2 OCR. We systematically evaluate their performance across multiple document quality levels, explicitly incorporating visual degradation and complexity reflective of real-world conditions. Furthermore, we introduce a comprehensive semantic answer evaluation benchmark that extends beyond traditional retrieval metrics, enabling a deeper understanding of each pipeline’s practical strengths and limitations.


We initially anticipated that the integrated visual-textual embedding of VLM-based approaches would offer superior robustness to visual noise, layout variations, and OCR-induced inaccuracies, potentially resulting in higher reliability for practical applications involving visually imperfect documents. However, our empirical findings indicate that VLM-based approaches struggle to generalize effectively to unseen data that they have not been explicitly fine-tuned on. 



\subsection{Contributions}

This study advances the state-of-the-art (SOTA) through three core contributions:
\begin{enumerate}
    \item 
    We present the first systematic empirical comparison between two leading paradigms for document retrieval augmentation: VLM-based RAG system, specifically ColQwen2 (7 billion parameters), and an advanced OCR-based RAG pipeline leveraging Llama 3.2 (90 billion parameters). To the best of our knowledge, previous evaluations have not explicitly addressed the substantial model-size disparity between lightweight VLM architectures commonly used in visual retrieval (e.g., ColQwen2) and significantly larger VLM for OCR-task.

    
    \item We introduce and execute a rigorous experimental protocol that systematically evaluates both systems across precisely controlled levels of document degradation. By manually categorizing  the documents based on their noise levels, we explicitly characterize each system’s robustness and performance under conditions found in real-world deployment scenarios. 

    \item We propose and validate a novel semantic evaluation benchmark explicitly designed for assessing end-to-end performance in knowledge-intensive document question-answering tasks. Unlike conventional evaluations focused solely on retrieval accuracy, our benchmark incorporates automated semantic metrics including Exact Match, BLEU, ROUGE-1, and ROUGE-L.

\end{enumerate}


\section{Related Work}
This section gives an overview of various document retrieval works related to this research, highlighting the limitations and strengths of differing approaches across both OCR and VLM retrieval methods. 

\subsection{Vision-Language Models for Document Understanding}
Early multimodal approaches to document understanding model both text and layout information. LayoutLM introduced unified transformer encoding textual tokens alongside spatial layout positions, greatly improving tasks such as form understanding and receipt parsing~\citep{xu2020layoutlm}. Further advancements incorporated visual features explicitly, such as DocFormer, which integrates text, layout, and visual embeddings within a transformer architecture~\citep{appalaraju2021docformer}. More recently, LayoutLMv3 extended this methodology by pre-training both text and image modalities with a unified masking strategy, achieving state-of-the-art results across text-heavy and image-intensive document tasks~\citep{huang2022layoutlmv3}. Another strong advancement, UDOP (Unifying Document Processing), introduced a generative vision-text-layout transformer model capable of different document AI tasks through a prompting-based framework, achieving state-of-the-art results across multiple document benchmarks~\citep{tang2023udop}. While powerful, these methods still depend on the extracted textual inputs from OCR, supplemented with visual information and layout context.

\subsection{OCR Integration with Retrieval-Augmented Generation}
RAG systems rely on the quality of input documents for effective retrieval. Liu et al. (2023) showed that increases in OCR noise significantly reduces retrieval accuracy in various tested RAG applications. Zhang et al. (2024) introduced OHRBench, a benchmark for evaluating how various levels of OCR-induced noise affect RAG pipelines \citep{zhang2024ocr}. In that work, Zhang et al. found that even state-of-the-art OCR systems often fail to construct high-quality knowledge bases for retrieval when the original documents contain what they call semantic and formatting noise.
Piryani et al. (2025) developed MultiOCR-QA, a multilingual dataset designed to test LLM performance on OCR-processed texts, in which the authors found that character-level OCR errors significantly reduce QA accuracy \citep{piryani2025multiocrqa}.

Transformer-based OCR models like Nougat and Llama 3.2 have improved the OCR bottleneck, but their effectiveness under real-world noise conditions is still limited. Understanding the interaction between OCR fidelity and retrieval outcomes is important for evaluating when traditional OCR pipelines are sufficient compared to newer approaches such as VLM-based retrieval (e.g., ColPali) that claim higher levels of robustness. This work situates itself within this space by directly comparing text embedding based RAG pipelines, driven by OCR, versus vision embedding RAG systems under varying document quality controls. 

\subsection{Vision-Language Models for Document Retrieval}
VLMs offer an alternative to OCR methods by embedding entire document images, which includes text, layout, and visual context into a shared representation space. This approach bypasses the traditional OCR step, which could potentially reduce issues arising from text extraction error.
CLIP \citep{radford2021learning} was one of the first large-scale VLMs to align visual and textual embeddings using contrastive learning. SigLIP \citep{zhai2023sigmoid}, modified CLIP by introducing a sigmoid-based loss function that allows for more stable training and better scalability across noisy or individual text-image pairs. These models laid the groundwork for more document-specific architectures such as PaliGemma \citep{beyer2024paligemma}, which combines SigLIP’s image encoder with Gemma’s decoder to handle token-level document retrieval tasks.

ColBERT (Contextualized Late Interaction - BERT), expands upon BERT by adding a late interaction mechanism. The main difference from BERT is that this mechanism delays the interaction between query and document until the final scoring stage \citep{khattab2020colbert}. Similarly to BERT, ColBERT also stores embeddings at the word-token level for both query and document, instead of a single embedding at the document level, and then afterward scores similarity between each combination of word-document embedding. This in turn allows for a more in-depth similarity calculation between query and document compared to its single-embedding per document counterpart, providing higher retrieval accuracy. ColPali \citep{faysse2024colpali} additionally builds on this approach by combining the PaliGemma model with ColBERT-style late interaction, to match visual document embeddings with token-level query embeddings.

\subsection{ColPali and the ViDoRe Benchmark}
ColPali is a RAG design that uses a VLM to bypass the traditional OCR-to-text pipeline by directly embedding image patches across the entire document \citep{faysse2024colpali}. By utilizing the late interaction mechanism seen in ColBERT, alongside PaliGemma's ability to generate multi-vector embeddings for document images (patches) that capture both the textual and visual data inside each document, ColPali is able to efficiently match queries with relevant documents. Overall, ColPali aims to increase the accuracy of document retrieval, while also reducing training time by embedding text and visual chunks simultaneously in a single, end-to-end method.
To evaluate performance, the authors of ColPali proposed the ViDoRe (Visual Document Retrieval) benchmark \citep{faysse2024colpali}. This dataset includes 127,346 high-quality PDF pages that generally all have well-structured layouts and legible content. Models are scored using metrics like NDCG@5, which tests the ability to retrieve the correct page given a specific query. While ColPali demonstrates strong performance on this benchmark, the controlled nature of the datasets —high resolution, consistent formatting—limits its generalizability. Additionally, in its OCR-to-text RAG baseline, ColPali uses Tesseract, which is not considered state-of-the-art.
Our work builds upon ViDoRe by testing whether ColPali’s retrieval advantages persist under more realistic and visually degraded conditions while also adding a more competitive OCR method (Llama 3.2 90B). This work also introduces a semantic evaluation step that compares the retrieved content’s answer to a ground truth answer, allowing the ability to move beyond exact page-match metrics and assess downstream utility in RAG pipelines.

\subsection{Comparative Benchmarks for Vision-Language Models}
Recent studies have introduced benchmarks to evaluate the strengths and weaknesses of VLMs across various document understanding tasks. HELM (Holistic Evaluation of Language Models) is a framework developed to provide a standardized, multidimensional evaluation of language models \citep{liang2022helm}. VHELM extends this framework to the vision-language setting \citep{lee2024vhelm}. LVLM-eHub provides a large-scale benchmark for evaluating instruction-tuned VLMs on multimodal tasks, to obtain a better understanding of generalization and issues with overfitting \citep{xu2024lvlm}.
Other evaluations additionally explore architectural comparisons and reliability. For example, Mamba-based models have shown better performance compared to traditional transformer-based VLMs in vision-language tasks \citep{waleffe2024mamba}. DeCC introduces a task decomposition framework to evaluate answer consistency and the robustness of VLMs \citep{yang2024decc}. Overall, these benchmarks highlight both the promise and limitations of VLMs in document-level reasoning, which is one of the primary focuses of this work on image-based retrieval.
In addition, this work complements these efforts by evaluating VLM-based retrieval approaches, specifically in the context of noisy, real-world documents and directly comparing them against OCR-based RAG pipelines.

\begin{figure}[!ht]
    \centering
    \includegraphics[width=0.8\linewidth]{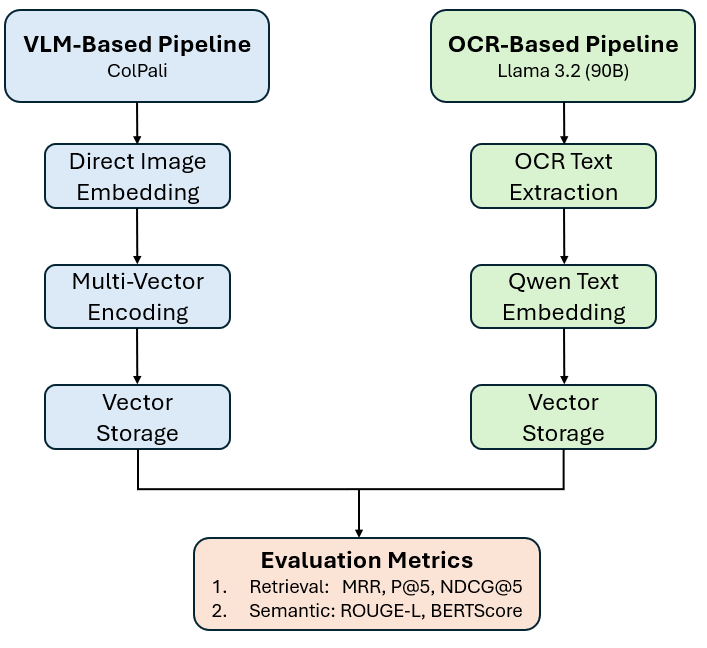}
    \caption{Overview of the experimental pipeline comparing VLM-based and OCR-based RAG systems.}
    \label{fig:overview}
\end{figure}
\section{Methodology}

In this section, we present the detailed methods and formal mathematical descriptions for the RAG pipelines compared in our study. We outline the VLM-based pipeline, the OCR-based pipeline, and the rigorous evaluation framework used to assess both retrieval and semantic accuracy. The overview of the proposed framework is shown in Figure~\ref{fig:overview}.

\subsection{System Architectures}

We rigorously evaluate two distinct RAG system paradigms: a VLM-based approach that directly embeds images, and an OCR-based pipeline that extracts text before embedding.

\subsubsection{VLM-Based RAG Pipeline}

Our vision-based RAG pipeline follows the ColPali architecture proposed by~\citet{faysse2024colpali}, specifically designed to bypass OCR preprocessing. The core architecture utilizes the PaliGemma model, integrating SigLIP-So400m as the image encoder and Gemma-2B as the textual encoder-decoder. Formally, a given document image \( D_i \) is partitioned into \( m \) non-overlapping patches \( \{p_{i,1}, p_{i,2}, \dots, p_{i,m}\} \). Each patch \( p_{i,j} \) is encoded by the VLM encoder \( f_{\text{VLM}} \) into embedding vectors:
\[
\mathbf{e}_{i,j}^{\text{(VLM)}} = f_{\text{VLM}}(p_{i,j}; \theta_{\text{VLM}}), \quad \mathbf{e}_{i,j}^{\text{(VLM)}} \in \mathbb{R}^d,
\]
where \( d \) denotes embedding dimensionality and \( \theta_{\text{VLM}} \) the model parameters. The resulting multi-vector embedding representation is:
\[
E_i^{\text{(VLM)}} = \{\mathbf{e}_{i,1}^{\text{(VLM)}}, \dots, \mathbf{e}_{i,m}^{\text{(VLM)}}\}.
\]

At retrieval time, queries \( q \) are encoded using a textual embedding model \( g \) parameterized by \( \phi \):
\[
\mathbf{q} = g(q; \phi), \quad \mathbf{q} \in \mathbb{R}^d.
\]

Retrieval scoring employs ColBERT-style late-interaction, computing similarity scores between query and document patch embeddings:
\[
s_i^{(\text{VLM})}(q, D_i) = \max_{j \in [1, m]} \frac{\mathbf{q}^\top \mathbf{e}_{i,j}^{\text{(VLM)}}}{\|\mathbf{q}\| \|\mathbf{e}_{i,j}^{\text{(VLM)}}\|}.
\]

At the time of our experiments, the SOTA model from the ColPali family was ColQwen2 (7B), which we employed for the experiments presented here.

\subsubsection{OCR-Based RAG Pipeline}

Our OCR-based pipeline leverages advanced OCR extraction combined with semantic-rich text embeddings. Document images \(D_i\) first undergo OCR via an OCR model \(h_{\text{OCR}}\):
\[
T_i = h_{\text{OCR}}(D_i),
\]
where \(T_i\) is the extracted textual content. This text is subsequently embedded using the Qwen embedding model \( g(\cdot; \phi) \) to produce a single, dense semantic embedding:
\[
\mathbf{e}_i^{\text{(OCR)}} = g(T_i; \phi), \quad \mathbf{e}_i^{\text{(OCR)}} \in \mathbb{R}^d.
\]

Documents are stored in a vector database based on these embeddings. Query embedding follows the same procedure, ensuring consistent semantic representation. Retrieval scores for OCR embeddings use cosine similarity:
\[
s_i^{(\text{OCR})}(q, D_i) = \frac{\mathbf{q}^\top \mathbf{e}_i^{\text{(OCR)}}}{\|\mathbf{q}\| \|\mathbf{e}_i^{\text{(OCR)}}\|}.
\]

This pipeline's accuracy directly depends on OCR quality, making it potentially sensitive to visual degradation and complex layouts.

\subsection{DocDeg Dataset}



We introduce \textbf{DocDeg}, a curated dataset specifically designed to evaluate Retrieval-Augmented Generation (RAG) systems under realistic visual degradation conditions. DocDeg comprises 4,196 diverse documents obtained from the U.S. Department of Energy’s Office of Scientific and Technical Information (OSTI) and includes a wide variety of document types, such as academic papers, technical reports, memos, presentation slides, and handwritten notes, spanning both scanned and digitally generated quality based on OSTI source~\footnote{https://www.osti.gov/}. The scanned documents often exhibit real-world noise, artifacts, and quality loss often encountered in practical RAG deployments.

Each document page was manually labeled by two expert annotators into four distinct degradation categories reflecting visual clarity: Level 0 (native digital documents with no degradation), Level 1 (high-quality scans with minimal visual noise or artifacts), Level 2 (imperfect scans exhibiting moderate blur, faint text, or minor artifacts), and Level 3 (severely degraded pages featuring significant distortion, extensive noise, and considerable content loss).

\subsubsection{Feature Annotation}

To enable a more detailed understanding of retrieval challenges, each document in the DocDeg dataset was manually annotated among a range of 12 structural and content features. Initially, 5,800 document pages were labeled by two annotators. After filtering out approximately 1,600 documents that lacked sufficient textual or visual information (e.g., blank pages, cover sheets, low-content slides), the final DocDeg dataset consists of 4,196 information-rich documents used in our experiments. Our analysis focuses on the degree of visual distortion or degradation. We judged quantified degradation on a scale of 0 to 3, with the following categories:

\begin{itemize}
    \item \textbf{Level 0 (Digital Copy):} Native digital document with no visible degradation.
    \item \textbf{Level 1 (Perfect Scan):} High-quality scanned document with minimal or no noise/artifacts.
    \item \textbf{Level 2 (Imperfect Scan):} Noticeable visual defects such as mild blurring, faded or missing text, or scanning artifacts.
    \item \textbf{Level 3 (Severely Degraded):} Document is almost illegible, with heavy noise, distortion, or severe content loss.
\end{itemize}

A zoomed in example of each degradation level is provided in Figure ~\ref{fig:subfigs2}. 


\subsubsection{Q\&A pair generation}

We constructed a rigorous evaluation benchmark by generating ten unique, nuanced question-answer pairs per document using Llama 3.3 (70B). We leveraged Sambanova Systems’ API calls to deploy the LLaMA mod-
els for Q\&A metadata synthesis and retrieval evaluation. These pairs were carefully designed to capture specific factual details from each document, explicitly avoiding direct verbatim copying from the source text. Leveraging these generated question-answer pairs, we performed two distinct evaluations of RAG system performance: (1) Document Retrieval Assessment, where each generated question was treated as a query to assess the system's ability to correctly retrieve the corresponding source document, and (2) Semantic Answer Generation Evaluation, where we evaluated whether the RAG-augmented language model could accurately produce the reference answer provided in the generated pairs when conditioned on retrieved document context. This dual-evaluation setup enabled a comprehensive assessment of both retrieval precision and downstream semantic accuracy of the RAG systems under investigation.

\subsection{Evaluation Framework}

To rigorously assess both retrieval effectiveness and downstream semantic accuracy, we employ a dual evaluation framework consisting of standard retrieval metrics and a novel semantic answer evaluation benchmark. 

\subsubsection{Retrieval Evaluation Metrics}

\begin{figure*}[ht]
  \centering
  \begin{subfigure}[b]{0.24\textwidth}
    \includegraphics[width=\linewidth, trim=500 500 500 500, clip]{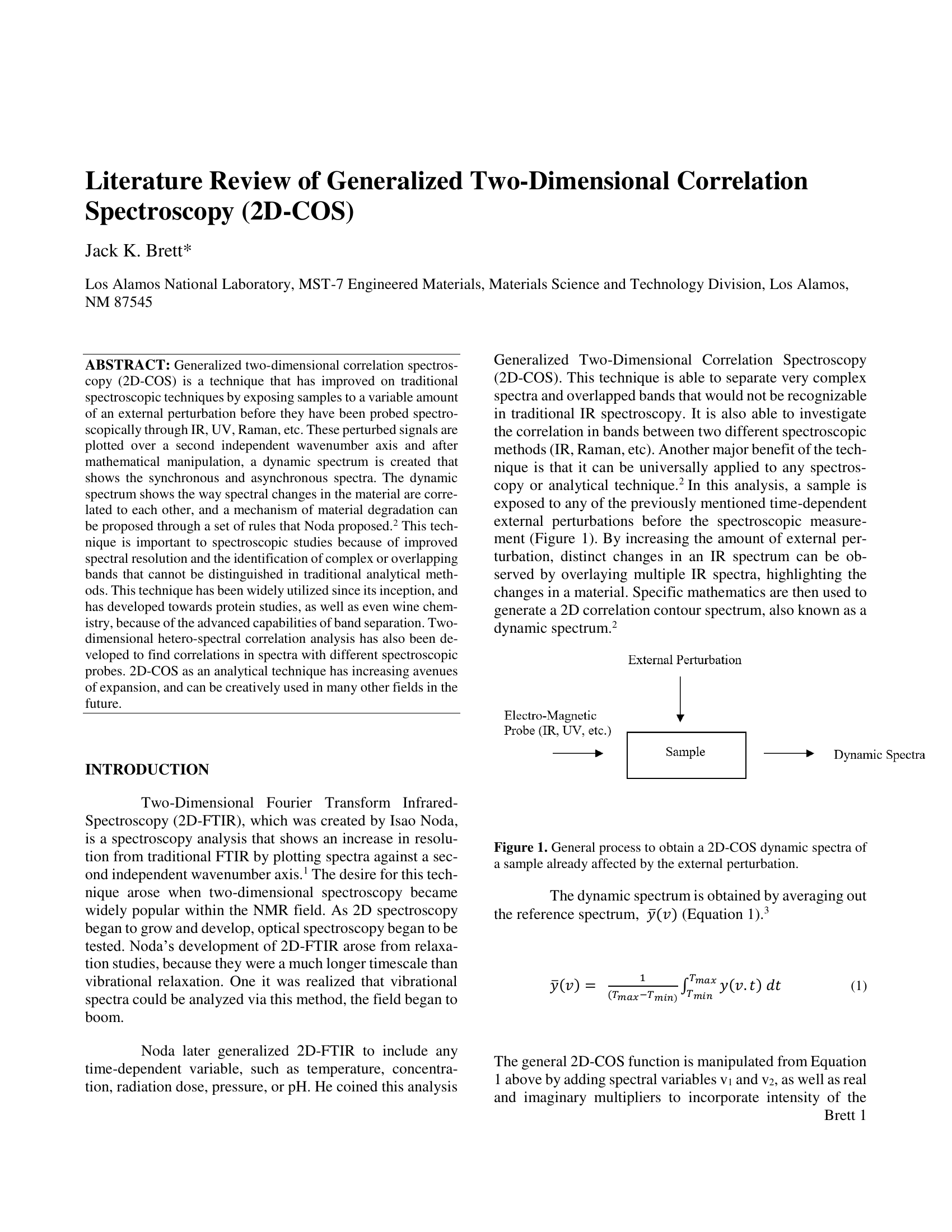}
    \caption{level-0}
  \end{subfigure}\hspace{0.5em}
  \begin{subfigure}[b]{0.24\textwidth}
    \includegraphics[width=\linewidth, trim=500 500 500 500, clip]{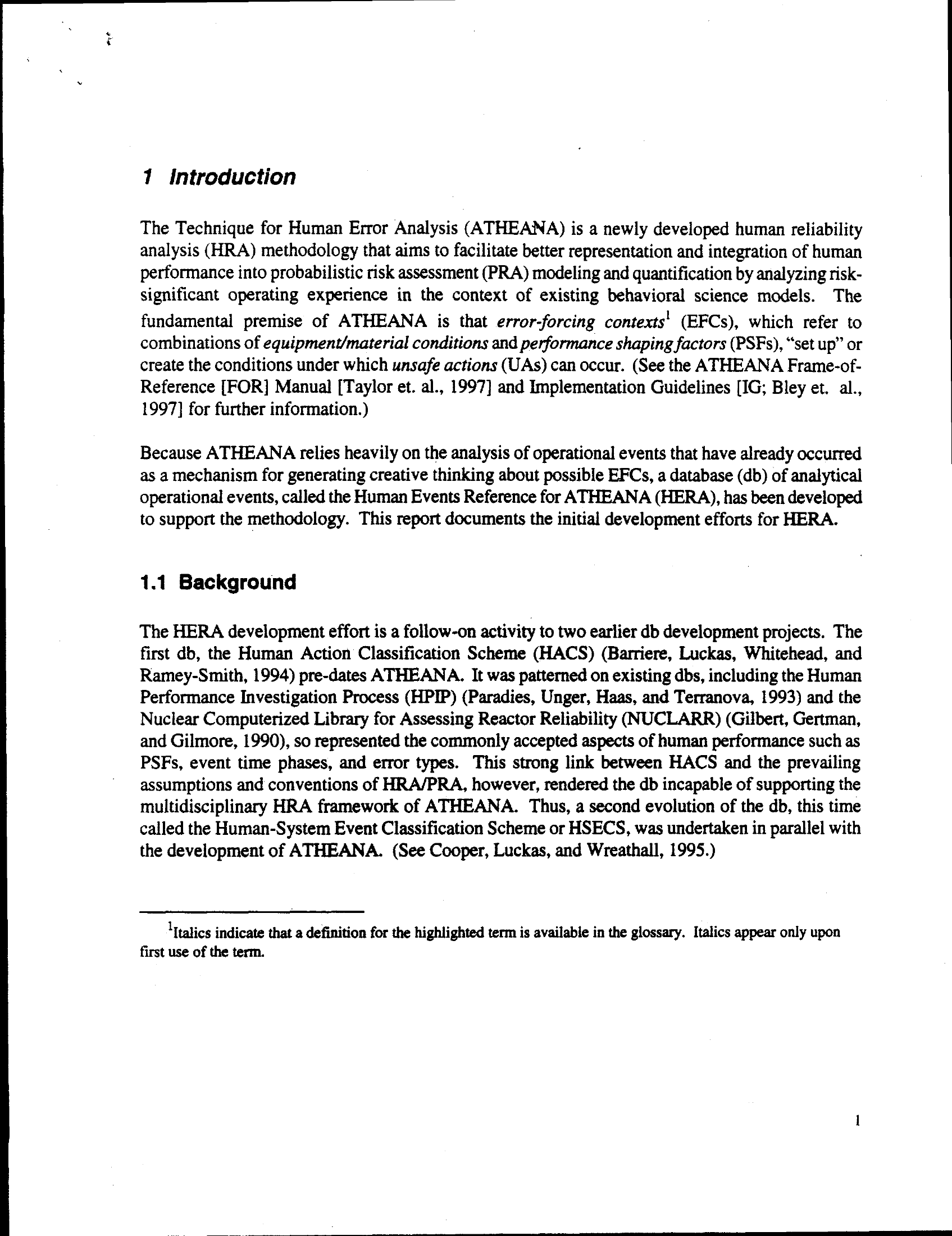}
    \caption{level-1}
  \end{subfigure}\hspace{0.5em}
  \begin{subfigure}[b]{0.24\textwidth}
    \includegraphics[width=\linewidth, trim=500 500 500 500, clip]{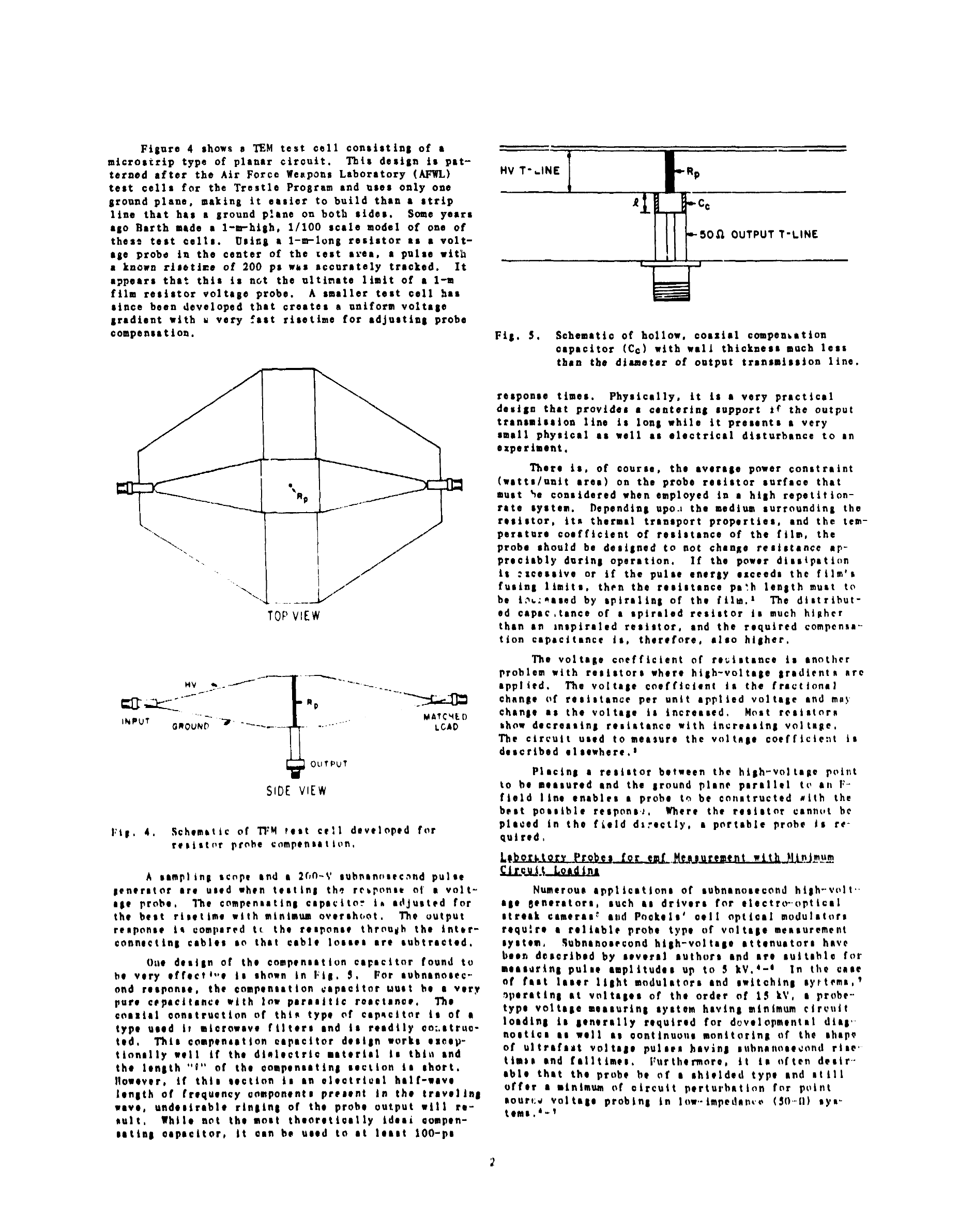}
    \caption{level-2}
  \end{subfigure}\hspace{0.5em}
  \begin{subfigure}[b]{0.24\textwidth}
    \includegraphics[width=\linewidth, trim=500 500 500 500, clip]{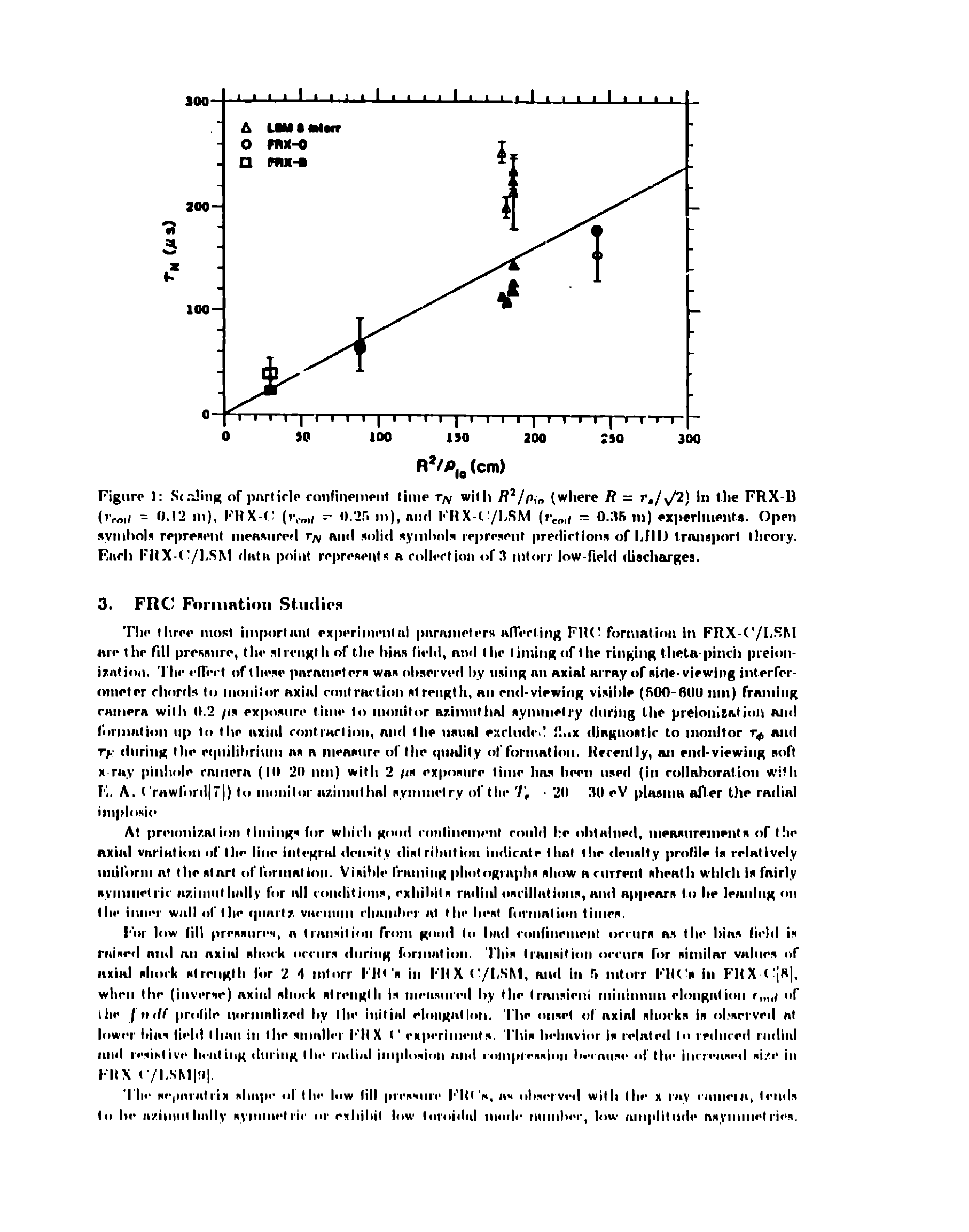}
    \caption{level-3}
  \end{subfigure}
  \caption{Zoomed-in view of documents with different degradation levels}
  \label{fig:subfigs2}
\end{figure*}

We utilize well-established retrieval metrics to quantitatively measure retrieval performance:

\begin{itemize}
    \item \textbf{Mean Reciprocal Rank (MRR)} evaluates retrieval quality by averaging the reciprocal rank positions of correct documents:
    \[
    \text{MRR} = \frac{1}{|Q|}\sum_{q \in Q}\frac{1}{r_q}.
    \]

    \item \textbf{Recall@k} measures the proportion of relevant documents retrieved among the top-k results:
    \[
    \text{Recall@k} = \frac{|\{\text{relevant documents in top } k\}|}{|\{\text{relevant documents total}\}|}.
    \]

    \item \textbf{Normalized Discounted Cumulative Gain (NDCG@k)} quantifies retrieval ranking quality by considering the position and relevance of documents retrieved:
    \[
    \text{NDCG@k} = \frac{\sum_{i=1}^k \frac{2^{rel_i}-1}{\log_2(i+1)}}{\sum_{i=1}^k \frac{2^{rel_i^{ideal}}-1}{\log_2(i+1)}}.
    \]
\end{itemize}

The degradation labels were used to segment the dataset for all retrieval evaluations reported in this paper. Figure ~\ref{fig:subfigs2} provides a zoomed in example for each of the four human evaluated text degradation levels. 

\subsubsection{Semantic Answer Evaluation}

In addition to retrieval metrics, we evaluate semantic accuracy using a robust end-to-end evaluation framework involving automated metrics for semantic comparison. We first create reference answers by generating ten unique question-answer pairs per document using Llama 3.3 (70B)~\citep{grattafiori2024llama3}. These pairs are designed to emphasize nuanced, factual details and avoid verbatim copying from source documents.

Semantic accuracy is evaluated using the following metrics:

\begin{itemize}
    \item \textbf{Exact Match (EM)} measures the strict accuracy by comparing generated answers \(a_i\) to references \(r_i\):
    \[
    \text{EM} = \frac{1}{N}\sum_{i=1}^N \mathbf{1}(a_i = r_i).
    \]

    \item \textbf{BLEU Score} evaluates lexical precision via n-gram overlap:
    \[
    \text{BLEU}(a, r) = BP \cdot \exp\left(\sum_{n=1}^{4} w_n\log p_n(a,r)\right).
    \]

    \item \textbf{ROUGE Scores (ROUGE-1, ROUGE-L)} measure lexical completeness and structural coherence. ROUGE-1 considers unigram overlap:
    \[
    \text{ROUGE-1} = \frac{|a \cap r|}{|r|},
    \]
    while ROUGE-L uses the longest common subsequence (LCS) overlap:
    \[
    \text{ROUGE-L} = \frac{(1+\beta^2)\text{LCS}(a,r)}{|a|+\beta^2|r|}, \quad \text{typically } \beta=1.
    \]
\end{itemize}

\subsection{Retrieval Performance}

RAG practitioners typically deploy systems "out of the box", relying on pre-trained models without additional fine-tuning on their specific document collections due to practical constraints such as cost, time, expertise and data availability. With that in mind, our primary evaluation focuses on how each pipeline performs under realistic, visually degraded conditions using pre-trained models without performing task-specific fine-tuning on the DocDeg dataset for either the VLM-based or the OCR-based retrieval components. This approach allows us to assess the generalizability and robustness of these architectures when faced with unfamiliar document types and degradation patterns, which is crucial for understanding their practical utility.


We report retrieval metrics—Mean Reciprocal Rank (MRR), Recall@5, and Normalized Discounted Cumulative Gain at rank 5 (NDCG@5)—across four levels of increasing document degradation on this dataset.

In addition to DocDeg, we also conducted supplementary experiments on the ViDoRe benchmark (specifically, the DocVQA subset). This evaluation compared three approaches:

\begin{itemize} \item \textbf{ColQwen (baseline)}: ColPali retrieval with their SOTA model, ColQwen2-v1, fine-tuned on the ViDoRe dataset.  \item \textbf{OCR+Qwen2 (Nougat)}: OCR extraction by rerunning the ViDoRe images through the Nougat OCR model, followed by Qwen2 7B Instruct, again without fine-tuning. 
\item \textbf{OCR+Qwen2 (Llama)}: OCR extraction by rerunning the ViDoRe images through the Llama3.2 90B OCR model, followed by Qwen2 7B Instruct, again without fine-tuning. 
\end{itemize}

\section{Results and Discussion} 
In this section, the findings from our experiments are presented, an analysis of performance differences is given, and the resulting implications are discussed in the context of document understanding and retrieval systems.
\begin{figure*}[!ht]
    \centering
    \includegraphics[width=0.8\linewidth]{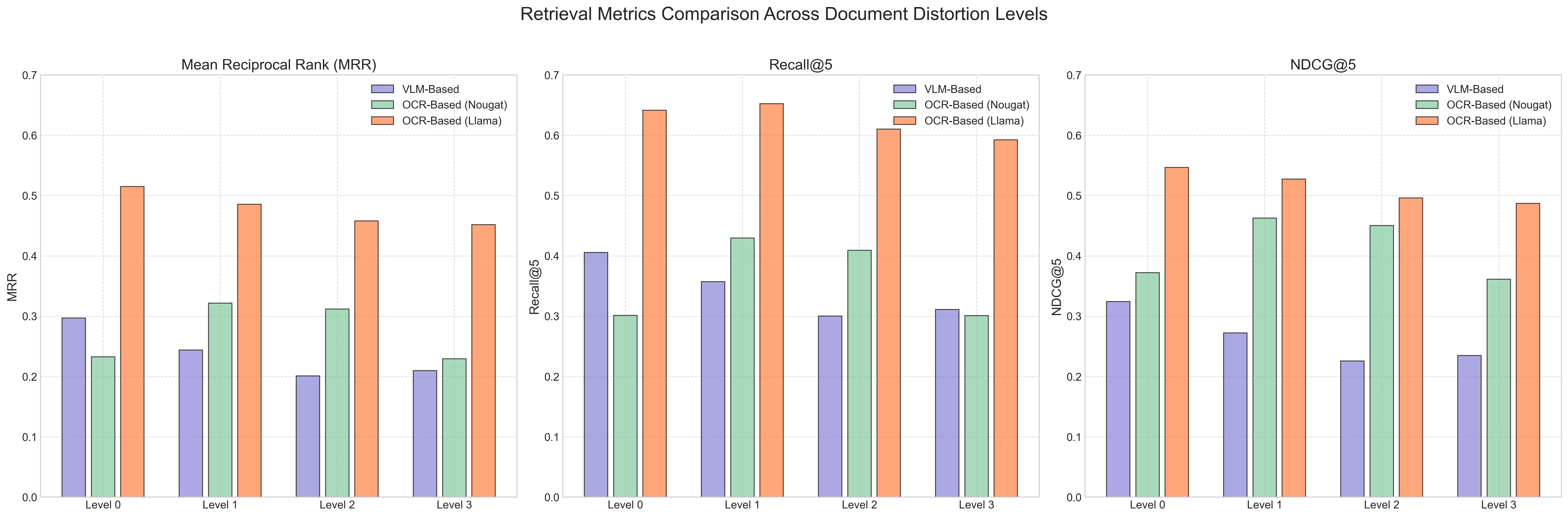}
    \caption{Retrieval performance across document quality levels}
    \label{fig:enter-label}
\end{figure*}

\subsection{Retrieval Performance Across Document Quality Levels}

To evaluate the robustness and accuracy of each pipeline across realistic conditions, we systematically assessed their retrieval performance over four incremental document degradation levels, ranging from pristine (Level 0) to severely degraded (Level 3). We report Mean Reciprocal Rank (MRR), Recall@5, and Normalized Discounted Cumulative Gain at rank 5 (NDCG@5). 

Table~\ref{tab:retrieval_combined} compares a vision-language model (VLM)-based pipeline against two OCR-based pipelines, one using nougat OCR and another using Llama 3.2 for OCR.

\begin{table}[h]
\centering
\small
\caption{
Retrieval performance metrics across document degradation levels.
\newline
\textit{VLM-Based} = Vision-Language Model pipeline,
\textit{OCR-Based (Nougat)} = OCR pipeline with Nougat OCR,
\textit{OCR-Based (Llama)} = OCR pipeline with Llama 3.2 OCR.
}
\label{tab:retrieval_combined}
\resizebox{\linewidth}{!}{%
\begin{tabular}{@{}lccccc@{}}
\toprule
\textbf{\shortstack{Distortion\\Level}} & \textbf{Questions} & \textbf{Metric} & \textbf{VLM-Based} & \textbf{\shortstack{OCR-Based\\(Nougat)}} & \textbf{\shortstack{OCR-Based\\(Llama)}} \\
\midrule
Level 0 & 11190 & MRR       & 0.2971 & 0.2327 & \textbf{0.5151} \\
        &       & Recall@5  & 0.4057 & 0.3013 & \textbf{0.6415} \\
        &       & NDCG@5    & 0.3242 & 0.2721 & \textbf{0.5468} \\
\midrule
Level 1 & 19154 & MRR       & 0.2440 & 0.3218 & \textbf{0.4857} \\
        &       & Recall@5  & 0.3573 & 0.4298 & \textbf{0.6525} \\
        &       & NDCG@5    & 0.2722 & 0.3627 & \textbf{0.5275} \\
\midrule
Level 2 & 6864  & MRR       & 0.2012 & 0.3119 & \textbf{0.4579} \\
        &       & Recall@5  & 0.3003 & 0.4093 & \textbf{0.6104} \\
        &       & NDCG@5    & 0.2259 & 0.3502 & \textbf{0.4961} \\
\midrule
Level 3 & 4748  & MRR       & 0.2098 & 0.2295 & \textbf{0.4520} \\
        &       & Recall@5  & 0.3112 & 0.3010 & \textbf{0.5925} \\
        &       & NDCG@5    & 0.2350 & 0.2612 & \textbf{0.4872} \\
\midrule
\textbf{Total} & \textbf{41956} & MRR     & 0.2471 & 0.2858 & \textbf{0.4852} \\
        &       & Recall@5  & 0.3554 & 0.3774 & \textbf{0.6359} \\
        &       & NDCG@5    & 0.2740 & 0.3212 & \textbf{0.5229} \\
\bottomrule
\end{tabular}
}
\end{table}

\begin{table}[h]
\centering
\small
\caption{
Retrieval performance metrics across document degradation levels (slides removed).
\newline
\textit{VLM-Based} = Vision-Language Model pipeline,
\textit{OCR-Based (Nougat)} = OCR pipeline with Nougat OCR,
\textit{OCR-Based (Llama)} = OCR pipeline with Llama 3.2 OCR.
}
\label{tab:retrieval_no_slides}
\resizebox{\linewidth}{!}{%
\begin{tabular}{@{}lccccc@{}}
\toprule
\textbf{\shortstack{Distortion\\Level}} & \textbf{Questions} & \textbf{Metric} & \textbf{VLM-Based} & \textbf{\shortstack{OCR-Based\\(Nougat)}} & \textbf{\shortstack{OCR-Based\\(Llama)}} \\
\midrule
Level 0 & 3784 & MRR       & 0.1905 & 0.2819 & \textbf{0.3769} \\
        &      & Recall@5  & 0.2826 & 0.3979 & \textbf{0.5322} \\
        &      & NDCG@5    & 0.2134 & 0.3109 & \textbf{0.4157} \\
\midrule
Level 1 & 16974 & MRR      & 0.2347 & 0.3416 & \textbf{0.4782} \\
        &       & Recall@5 & 0.3476 & 0.4593 & \textbf{0.6502} \\
        &       & NDCG@5   & 0.2628 & 0.3711 & \textbf{0.5213} \\
\midrule
Level 2 & 6794  & MRR      & 0.2005 & 0.3139 & \textbf{0.4557} \\
        &       & Recall@5 & 0.2997 & 0.4122 & \textbf{0.6086} \\
        &       & NDCG@5   & 0.2252 & 0.3386 & \textbf{0.4941} \\
\midrule
Level 3 & 4748  & MRR      & 0.2098 & 0.2295 & \textbf{0.4520} \\
        &       & Recall@5 & 0.3112 & 0.3010 & \textbf{0.5925} \\
        &       & NDCG@5   & 0.2997 & 0.2474 & \textbf{0.4872} \\
\midrule
\textbf{Total} & \textbf{32300} & MRR    & 0.2186 & 0.3123 & \textbf{0.4578} \\
        &       & Recall@5 & 0.3244 & 0.4189 & \textbf{0.6192} \\
        &       & NDCG@5   & 0.2449 & 0.3390 & \textbf{0.4982} \\
\bottomrule
\end{tabular}
}
\end{table}

As illustrated in Table~\ref{tab:retrieval_combined} and Figure~\ref{fig:enter-label}, the Llama OCR-Based pipeline consistently outperformed the VLM-Based pipeline on retrieval accuracy across all degradation levels evaluated on the DocDeg dataset for all metrics. Both systems show declines in accuracy as document quality decreases, though the OCR-Based scores still significantly outperformed the VLM-Based ones even under severe degradation. 

Comparing Nougat to Llama further highlights that the specific OCR engine used within the OCR-Based pipeline has a substantial impact on overall retrieval performance. The Llama 3.2 OCR engine consistently achieved higher retrieval metrics than Nougat across all distortion levels. This underscores the critical role that OCR quality plays in downstream retrieval tasks, particularly when deploying OCR-to-RAG systems in realistic, noisy document environments.

Nougat outperformed the VLM-based RAG pipeline on levels one through three and on the weighted average. Interestingly, Nougat performed worse on the highest quality documents. ~\ref{tab:retrieval_combined} shows retrieval results with slideshows removed from the dataset. Both Llama and ColQwen saw a reduction in retrieval accuracy on level zero documents when slides were excluded. Meanwhile, Nougat's performance on the slideshow-free data improved. 

Although ColPali retrieves the top-5 visually relevant documents based on query-image similarity, it does not natively support answer generation. To use ColPali in a full RAG pipeline, an additional step is required: either (1) apply OCR to the retrieved document images and pass the extracted text into a language model, or (2) use a vision-language model capable of question answering over images. Both options introduce additional computational cost, either in the form of OCR latency or VLM inference time. For this study’s semantic answer evaluation, we used the OCR text from the images retrieved by ColPali as context for the LLM, in order to enable a fair comparison across pipelines. 


\subsection{Semantic Answer Accuracy}

In addition to retrieval accuracy, we provide a comprehensive evaluation of the pipelines' end-to-end performance in question-answering scenarios using Exact Match, BLEU, ROUGE-1, and ROUGE-L (Table~\ref{tab:semantic_quality}). In addition to automated metrics, a small-scale manual review of approximately 40 generated question-answer pairs was conducted for qualitative verification. While not exhaustive, this review confirmed the quality and relevance of the question-answer pairs to the documents. 

\begin{table}[h]
\centering
\small
\caption{
Semantic answer quality across document degradation levels. 
\newline
\textit{VLM-Based} = ColQwen2, 
\textit{OCR-Based} = Llama 3.2 90B for OCR + Qwen2.
}
\label{tab:semantic_quality}
\resizebox{\linewidth}{!}{%
\begin{tabular}{@{}lccccc@{}}
\toprule
\textbf{Distortion Level} & \textbf{Questions} & \textbf{Metric} & \textbf{VLM-Based} & \textbf{OCR-Based} \\
\midrule
Level 0 & 11190 & Exact Match     & 0.0046 & \textbf{0.0080} \\
        &       & Average BLEU    & 0.0589 & \textbf{0.0753} \\
        &       & Average ROUGE-1 & 0.2088 & \textbf{0.2547} \\
        &       & Average ROUGE-L & 0.1912 & \textbf{0.2370} \\
\midrule
Level 1 & 19154 & Exact Match     & 0.0038 & \textbf{0.0066} \\
        &       & Average BLEU    & 0.0628 & \textbf{0.0804} \\
        &       & Average ROUGE-1 & 0.2167 & \textbf{0.2643} \\
        &       & Average ROUGE-L & 0.1960 & \textbf{0.2430} \\
\midrule
Level 2 & 6864  & Exact Match     & 0.0033 & \textbf{0.0057} \\
        &       & Average BLEU    & 0.0587 & \textbf{0.0751} \\
        &       & Average ROUGE-1 & 0.1975 & \textbf{0.2410} \\
        &       & Average ROUGE-L & 0.1785 & \textbf{0.2213} \\
\midrule
Level 3 & 4748  & Exact Match     & 0.0038 & \textbf{0.0066} \\
        &       & Average BLEU    & 0.0595 & \textbf{0.0762} \\
        &       & Average ROUGE-1 & 0.2044 & \textbf{0.2493} \\
        &       & Average ROUGE-L & 0.1826 & \textbf{0.2263} \\
\midrule
\textbf{Total} & \textbf{41956} & Exact Match     & 0.0039 & \textbf{0.0068} \\
        &        & Average BLEU    & 0.0607 & \textbf{0.0777} \\
        &        & Average ROUGE-1 & 0.2100 & \textbf{0.2556} \\
        &        & Average ROUGE-L & 0.1903 & \textbf{0.2355} \\
\bottomrule
\end{tabular}
}
\end{table}

Results from Table~\ref{tab:semantic_quality} show the semantic answer evaluation. Consistent with the results obtained from retrieval, the OCR-Based pipeline obtained higher scores when compared to the VLM-Based pipeline on every tested metric and degradation level. A notable difference between these results and the retrieval results, is that there does not seem to be an obvious decrease in score as degradation level increases. Overall, the results suggest that the potential benefits of the VLM's ability to handle visual nuance did not translate to improved semantic performance when compared to the superior results from the OCR-Based pipeline.

\subsection{Computational Efficiency Analysis}

A comparative analysis of indexing time, retrieval latency, and memory consumption for both pipelines is summarized in Table~\ref{tab:computational_efficiency}. The results reveal that the VLM-based pipeline is more efficient in both embedding generation and retrieval latency compared to the OCR-based pipeline. The VLM-based system achieved lower embedding times (0.4739s per document compared to 0.5503s) and substantially reduced retrieval latency (0.0010s per query vs. 0.0311s per query). Additionally, memory consumption per 1,000 documents was notably lower for the VLM-based pipeline (1.38 GB) relative to the OCR-based pipeline (9.5 GB). This substantial memory efficiency can be primarily attributed to the avoidance of storing extensive OCR-derived textual embeddings.

All experiments not involving Llama-based models were conducted on a node equipped with two NVIDIA A100 GPUs (40GB VRAM each), an AMD EPYC 7502 CPU, and 251 GB of system memory. However, the SentenceTransformer model was executed on a single A100 throughout. The second GPU remained unused. For OCR extraction and question-answer pair generation involving Llama 3.2 and Llama 3.3 models, we utilized a SambaNova SN40L platform with integrated RDU accelerators. Retrieval latency, embedding times, and memory usage metrics reported in Table~\ref{tab:computational_efficiency} reflect performance measured on these respective systems. Embedding of text and images was done with a batch size of 1. 

\begin{table}[h]
\centering
\caption{Computational Efficiency Comparison}
\label{tab:computational_efficiency}
\resizebox{\linewidth}{!}{%
\begin{tabular}{@{}lccc@{}}
\toprule
\textbf{Metric} & \textbf{VLM-Based} & \textbf{OCR-Based} \\
\midrule
OCR Time (per document) & N/A & 6s \\
Embedding Time (per document) & .4739s & .5503s \\
Retrieval Latency (per query) & 0.04252s & .0311s \\
Memory Usage (per 1k documents) & 1.38 GB & 9.5 GB \\
\bottomrule
\end{tabular}
}
\end{table}

\begin{figure*}[!ht]
    \centering
    \includegraphics[width=.6\linewidth]{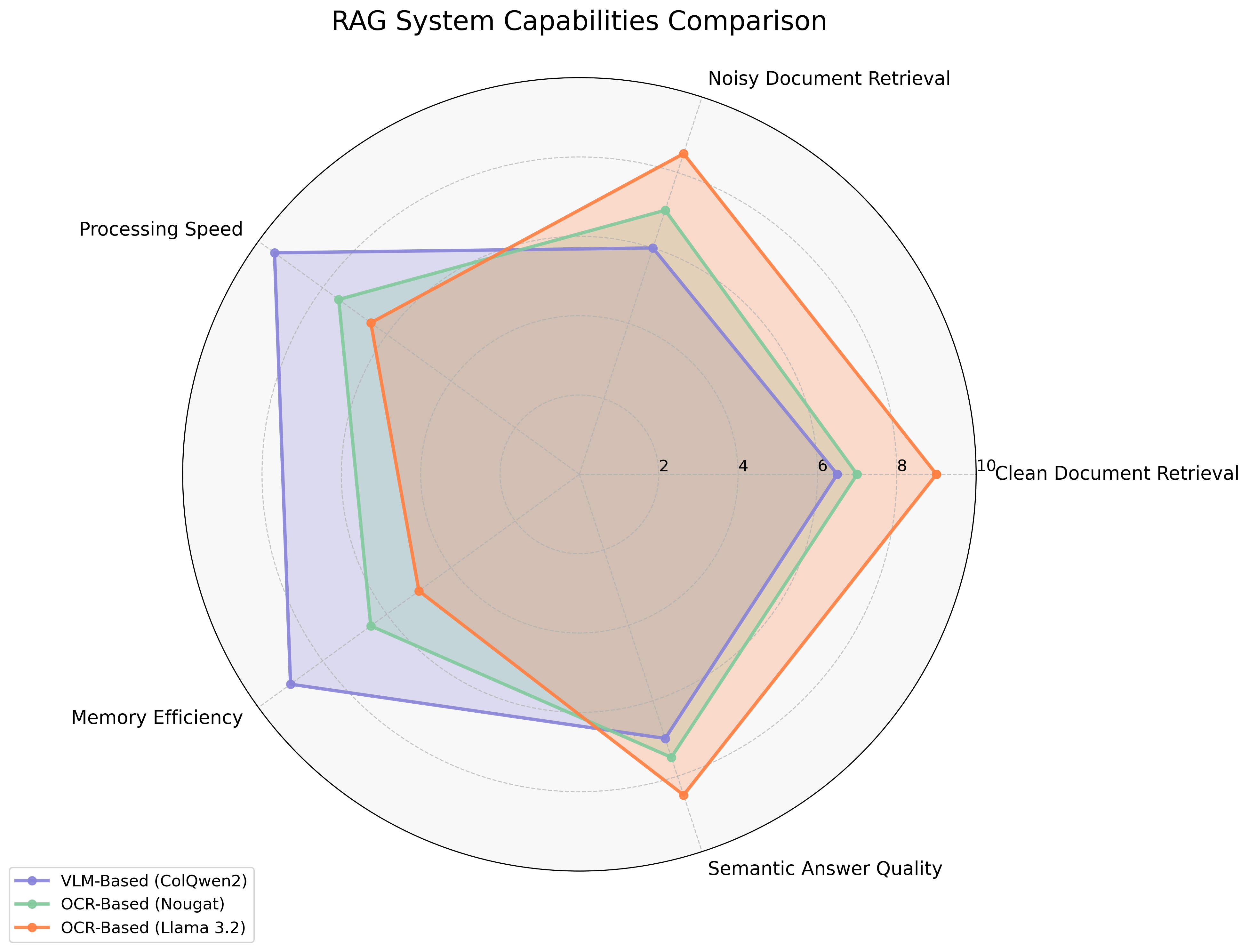}
    \caption{Comparative Analysis of RAG System Capabilities}
    \label{fig:radar}
\end{figure*}


\subsection{Overall assessment on DocDeg Dataset}
We leverage a RADAR plot to showcase overall performance of VLM vs OCR RAG shown in Figure~\ref{fig:radar}. Each axis is a linearly rescaled version of a concrete, published metric—chosen so that a score of 10 corresponds to the best value observed across \emph{all} pipelines and degradation levels, and a score of 0 would correspond to a value at least two standard deviations worse than the worst observed.\footnote{Because no pipeline actually reached that extreme, the lowest displayed score is $\approx5.5$.}  Specifically:
\begin{enumerate}
  \item \textbf{Clean–document retrieval} = $10\!\times\!\frac{\text{MRR}_{d=0}-\mu_{\text{MRR}}}{\sigma_{\text{MRR}}}$, clipped to $[0,10]$.
  \item \textbf{Noisy–document retrieval} = $10\!\times\!\frac{\text{MRR}_{d=3}-\mu_{\text{MRR}}}{\sigma_{\text{MRR}}}$.
  \item \textbf{Semantic answer quality} = $10\!\times\!\frac{\text{ROUGE--L}_{\text{total}}-\mu_{\text{R}}}{\sigma_{\text{R}}}$, where $\mu_{\text{R}},\sigma_{\text{R}}$ are the mean and s.d.\ of ROUGE-L across all runs.
  \item \textbf{Processing speed} = $10\!\times\!\frac{\min(T_{\text{ret}})}{T_{\text{ret}}}$, where $T_{\text{ret}}$ is per-query latency; lower is better.
  \item \textbf{Memory efficiency} = $10\!\times\!\frac{\min(M)}{M}$, where $M$ is memory per 1k documents.
\end{enumerate}
All raw values (\textit{MRR}, ROUGE-L, latency, memory) are reported in Tables \ref{tab:retrieval_combined} and \ref{tab:computational_efficiency}.  The resulting normalised scores for VLM, OCR+Nougat, and OCR+Llama are plotted without additional smoothing.  This transparent mapping ensures that the radar visualisation is a faithful, monotonic transformation of the quantitative results.

The radar plot  provides a comprehensive visualization of the relative strengths and weaknesses between VLM-based and OCR-based RAG systems across five critical dimensions. The VLM-based approach demonstrates exceptional performance in processing speed (9.5/10) and memory efficiency (9.0/10), reflecting its significantly faster retrieval latency (0.001s vs 0.031s) and lower memory footprint (1.38GB vs 9.5GB per 1k documents). However, it shows limitations in both clean document retrieval (6.5/10) and noisy document retrieval (6.0/10). In contrast, the OCR-based pipeline with Llama 3.2 exhibits superior performance in clean document retrieval (9.0/10), noisy document retrieval (8.5/10), and semantic answer quality (8.5/10), at the cost of greater computational demands. The Nougat OCR-based system offers a middle ground across most metrics. These findings reveal a clear trade-off: while modern OCR-based pipelines leveraging advanced models like Llama 3.2 provide better retrieval accuracy and semantic quality in realistic document settings, VLM-based approaches offer significant advantages in computational efficiency and resource utilization.

\subsection{ViDoRe Evaluation Results}

To further evaluate retrieval performance under cleaner document conditions, we conducted supplementary experiments on the DocVQA subset of the ViDoRe benchmark. We compared three retrieval pipelines: (1) the ColQwen2 model, representing the ViDoRe authors' best fine-tuned vision-language system based on ColPali; (2) an OCR-based pipeline using Alibaba-NLP/gte-Qwen2-7B-instruct with Nougat OCR outputs; and (3) the same OCR-based pipeline but preprocessing the images into text with the more advanced Llama 3.2 90B vision model.

Table~\ref{tab:vidore_results} summarizes the key retrieval metrics across the three approaches.

\begin{table}[h]
\centering
\caption{ViDoRe DocVQA Retrieval Comparison}
\label{tab:vidore_results}
\resizebox{\linewidth}{!}{%
\begin{tabular}{lccc}
\toprule
\textbf{\shortstack{Metric\\(Top-5)}} & \textbf{ColQwen2} & \textbf{\shortstack{OCR+Qwen2\\(Nougat)}} & \textbf{\shortstack{OCR+Qwen2\\(Llama)}} \\
\midrule
NDCG@5   & 0.6027 & 0.3373 & 0.3373\\
MAP@5    & 0.5813 & 0.3147 & 0.3147\\
MRR@5    & 0.5851 & 0.3164 & 0.3164\\
Recall@5 & 0.6674 & 0.4058 & 0.4058\\
\bottomrule
\end{tabular}
}
\end{table}

ColQwen2, which was fine-tuned on DocVQA, achieves the strongest retrieval performance, with an NDCG@5 of 0.6027 and a Recall@5 of 0.6674. 

Although the pages in ViDoRe are noise free, the benchmark is challenging because its queries are only loosely connected to the page text—typically short key phrases rather than full natural-language questions. Off-the-shelf retrieval models, whether vision-based or OCR-based, therefore struggle not with image quality but with recognising this subtle semantic link between a brief phrase and an entire page.

ColQwen2 succeeds because it has been fine-tuned on ViDoRe’s own query–page pairs, so its embeddings have learned to notice exactly those idiosyncratic links. In contrast, our Nougat- and Llama-OCR pipelines rely on a general-purpose text encoder that has never seen ViDoRe queries; they fall behind simply because their embeddings are semantically mis-aligned, not because their OCR output is faulty. Indeed, Nougat and Llama 3.2 produce almost identical retrieval scores, confirming that on such clean PDFs OCR quality is no longer the bottleneck—semantic alignment between the query and the page embeddings is.

\section{Conclusion and Future Work}

Our study rigorously compared OCR-based and VLM-based RAG pipelines, evaluating their performance and robustness across real degraded document scenarios. Our results show that, without task-specific fine-tuning, OCR-based approaches offer improved retrieval and generation performance in all evaluated settings, with some added cost at indexing time but more efficient performance at query time.

While OCR-based pipelines incur some additional latency during document ingestion, we found that they are actually faster at query time compared to VLM-based pipelines like ColPali. ColPali introduces higher query-time latency, even when using the same query encoder as the OCR pipeline. Furthermore, ColPali does not support end-to-end question answering directly: once documents are retrieved as images, either OCR or a vision-language QA model must be applied to generate a final answer, both of which add additional processing overhead.

Scaling to larger and more powerful vision encoders could potentially improve retrieval performance for VLM-based systems, but this would likely further increase both embedding time and query latency. Future research should compare fine-tuned OCR-based RAG pipelines with fine-tuned VLM-based RAG pipelines with a vision-language QA model to determine whether further performance improvements can be achieved.

\section{Acknowledgements}

This manuscript has been approved for unlimited release and has been assigned LA-UR-25-24280. This research was funded by the LANL ASC grant
AI4Coding and the LANL Institutional Computing Program, supported by the U.S. DOE NNSA under Contract No. 89233218CNA000001. We also thank Sambanova Systems for providing access to API calls for LLM inference utilized in this work

\bibliographystyle{ACM-Reference-Format}
\bibliography{main}

\end{document}